# Artificial and Biological Intelligence

Subhash Kak
Donald C. & Elaine T. Delaune Distinguished Professor of Electrical Engineering
Louisiana State University, Baton Rouge, LA 70803

In the early nineties there was much hope that neural networks would provide the next breakthrough in understanding biological intelligence, but that was belied by subsequent research. The recent success of several teams in meeting the $2 million DARPA Grand Challenge 2005 of designing an autonomous car that finishes a designated route of 175 mile over desert terrain featuring natural and man-made obstacles within 10 hours [1] raises the question if AI might be poised for another period of high support and increased expectations.

The quest for AI is also the subtext to debates outside of the field of computer science. Physics, for example, is the discovery of formal structures in nature, and each of these formal systems could be interpreted as a natural machine. The claim of some physicists that the universe itself is a giant machine is taken to complement the belief that true machine intelligence and self-awareness should arise after machine complexity has crossed a critical threshold.

But this leads to certain difficulties. Since machines only follow instructions, it is not credible that they should suddenly, on account of a greater number of connections between computing units, become endowed with self-awareness. On the other hand, if one accepts that machines will never become self-aware, one may ask why is the brain-machine conscious, whereas the silicon-computer is not? Perhaps the answer to this puzzle is that the brain is a self-organizing system which responds to the nature and quality of its interaction with the environment, whereas computers don't do that. But other ecological systems, which are biological communities that have complex interrelationship amongst its components, are self-organizing, without being self-aware. This suggests that while self-organization is a necessary pre-requisite for consciousness, it is not sufficient.

Yet another possibility is that current science, even when it considers self-organization and special structures of the brain, does not capture the essence of consciousness. The scientific framework may be incomplete in a variety of ways. We may not yet have discovered all the laws of nature, and our current theories may need major revision that has implications for our understanding of consciousness.

In truth, objective knowledge consists of many paradoxes. Accumulation of knowledge often amounts to making *ad hoc* choices in the underlying formal framework to conform to experience. The most fundamental, and a very ancient, antinomy is that between determinism and free will. Formal knowledge can at best be compared to a patchwork. The riddle is: How is a part of the physical world generating the individual's mental picture, in turn, creates the scientific theory that is able to describe nature so well? Why is mathematics so unreasonably effective?

Cognitive scientists and biologists have considered evolutionary aspects related to cognitive capacity, where consciousness is viewed as emerging out of language. Linguistic research on chimpanzees and bonobos has revealed that although they can be taught basic vocabulary of several hundred words, this linguistic ability does not extend to syntax. By contrast, small children acquire much larger vocabularies -- and use the words far more creatively -- with no overt training, suggesting that language is an innate capacity.

According to the nativist view, language ability is rooted in the biology of the brain, and our ability to use grammar and syntax is an instinct, dependent on specific modules of the brain. Therefore, we learn language as a consequence of a unique biological adaptation, and not because it is an emergent response to the problem of communication confronted by ourselves and our ancestors.

It has been suggested that human language capacities arose out of biological natural selection because they fulfill two clear criteria: an extremely complex and rich design and the absence of alternative processes capable of explaining such complexity. Other theories look at music and language arising out of sexual selection. But, howsoever imaginative and suggestive these models might be, they do not address the question of how the capacity to visualize models of world that are essential to language and consciousness first arise.

Finally, there is a philosophical critique of the search of a theory of consciousness. According to this critique, all that *normal* science can hope to achieve is a description of objects. But consciousness is a property of the subject, the experiencing "I", which, owing to its nature, for ever lies outside the pale of *normal* science. The experimenter cannot turn his gaze upon himself, and ordinary reality must have a dual aspect. This duality means that the world of objective causality is incomplete, creating a fundamental paradox: If objects are described by normal science, why is that science not rich enough to describe the body associated with the experiencing subject?

In recent work [2], I have considered evidence that negates the view that the brain is an *ordinary* machine. I argue that even with self-organization and hitherto-unknown quantum characteristics one cannot explain the capacities associated with the brain. A summary of these arguments follows.

**Brain and Mind**

The question of consciousness is connected to the relationship between brain and mind. Reductionism takes it that they are identical, and mind is only the sum total of the activity in the brain, viewed at a suitable higher level of representation. Opposed to this is the viewpoint that although mind requires a physical structure, it ends up transcending that structure.

The mind processes signals coming into the brain to obtain its understandings in the domains of seeing, hearing, touching, and tasting using its store of memories. But a cognitive act is an active process where the selectivity of the sensors and the accompanying processing in the brain is organized based on the expectation of the cognitive task and on effort, will and intention. Intelligence is a result of the workings of numerous active cognitive agents.

The reductionist approach to artificial intelligence emerged out of an attempt to mechanize logic in the 1930s. In turn, AI and computer science influenced research in psychology and neuroscience and the view developed that a cognitive act is a logical computation. This appeared reasonable as long as classical computing was the only model of effective computation. But with the advent of quantum computing theory, we know that the mechanistic model of computing does not capture all the power of natural computation.

Schrödinger spoke of the *arithmetic paradox* related to the mind as being "the *many* conscious egos from whose mental experiences the *one* world is concocted." He added [3] that there are only two ways out of the number paradox. "One way out is the multiplication of the world in Leibniz's fearful doctrine of monads: every monad to be a world by itself, no communication between them; the monad 'has no windows', it is 'incommunicado'. That none the less they all agree with each other is called 'pre-established harmony'. I think there are few to whom this suggestion appeals, nay who would consider it a mitigation at all of the numerical antinomy. There is obviously only one alternative, namely the unification of minds or consciousnesses."

**Languages of Description**

Progress in science is reflected in a corresponding development of language. The vistas opened up by the microscope, the telescope, tomography and other sensing devices have resulted in the naming of new entities and processes.

The language for the description of the mind in scientific discourse has not kept pace with the developments in the physical sciences. The mainstream discussion has moved from the earlier dualistic models of common belief to one based on the emergence of mind from the complexity of the parallel computer-like brain processes. The two old paradigms of determinism and autonomy, expressed sometimes in terms of separation and interconnectedness, show up in various guises. Which of the two of these is in favor depends on the field of research and the prevailing fashions. Although quantum theory has provided the foundation for physical sciences for seventy years, it is only recently that holistic, quantum-like operations in the brain have been considered. This fresh look has been prompted by the setbacks suffered by the various artificial intelligence projects and also by new analysis and experimental findings.

The languages used to describe the workings of the brain have been modeled after the dominant scientific paradigm of the age. The rise of mechanistic science saw the conceptualization of the mind as a machine. Although the neural network approach has had considerable success in modeling many counterintuitive illusions, there exist other processes in human and nonhuman cognition that appear to fall outside the scope of such models. Briefly, the classical neural network model does not provide a resolution to the question of binding of patterns: How do the neuron firings in the brain come to have specific meanings or lead to specific images?

Considering that the physical world is described at its most basic level by quantum mechanics, how can classical computational basis underlie the description of the structure (mind) that in turn is able to comprehend the universe? How can machines, based on classical logic, mimic biological computing? One may argue that ultimately the foundation on which the circuitry of classical computers is based is at its deepest level described by quantum mechanics. Nevertheless, actual computations are governed by a binary logic which is very different from the tangled computations of quantum mechanics. And since the applicability of quantum mechanics is not constrained, in principle, by size or scale, classical computers do appear to be limited.

Why cannot a classical computer reorganize itself in response to inputs? If it did, it will soon reach an organizational state associated with some energy minimum and will then stop responding to the environment. Once this state has been reached the computer will now merely transform data according to its program. In other words, a classical computer does not have the capability to be selective about its inputs. This is precisely what biological systems can do with ease.

Most proposals on considering brain function to have a quantum basis have done so by default. In short the argument is: There appears to be no resolution to the problem of the binding of patterns and there are non-local aspects to cognition; quantum behavior has non-local characteristics; so brain behavior might have a quantum basis.

Newer analysis has led to the understanding that one needs to consider reorganization as a primary process in the brain--- this allows the brain to define the context. The signal flows now represent the processing or recognition done within the reorganized hardware. Such a change in perspective can have significant implications. Dual signaling schemes eventually need an explanation in terms of a binding field; they do not solve the basic binding problem in themselves but they do make it easier to understand the process of adaptation.

**Biological Intelligence**

For all computational models, the question of the emergence of intelligence is a basic one. Solving a specified problem, that often requires searching or generalization, is taken to be a sign of AI, which is assumed to have an all or none quality. But biological intelligence has gradation. Animal performance depends crucially on its normal behavior. It may be argued that all animals are *sufficiently* intelligent because they survive in their ecological environment. Nevertheless, even in cognitive tasks of the kind normally associated with human intelligence, animals may perform well. Thus rats might find their way through a maze, or dolphins may solve logical problems to or problems involving some kind of generalization. These performances could, in principle, be used to define a gradation.

It has generally been assumed that the tasks that set the human apart from the machine are those that relate to abstract conceptualization best represented by language understanding. But nobody will deny that deaf-mutes, who don't have a language, do think. Language is best understood as a subset of a large repertoire of behavior. Research has now established that animals think and are capable of learning and problem solving.

Since nonhumans do not use abstract language, their thinking is based on discrimination at a variety of levels. If such conceptualization is seen as a result of evolution, it is not necessary that this would have developed in exactly the same manner for all species. Other animals learn concepts nonverbally, so it is hard for humans, as verbal animals, to determine their concepts. It is for this reason that the pigeon has become a favorite with intelligence tests; like humans, it has a highly developed visual system, and we are therefore likely to employ similar cognitive categories. It is to be noted that pigeons and other animals are made to respond in extremely unnatural conditions in Skinner boxes of various kinds. The abilities elicited in research must be taken to be merely suggestive of the intelligence of the animal, and not the limits of it.

In a classic experiment, Herrnstein [4] presented 80 photographic slides of natural scenes to pigeons who were accustomed to pecking at a switch for brief access to feed. The scenes were comparable but half contained trees and the rest did not. The tree photographs had full views of single and multiple trees as well as obscure and distant views of a variety of types. The slides were shown in no particular order and the pigeons were rewarded with food if they pecked at the switch in response to a tree slide; otherwise nothing was done. Even before all the slides had been shown the pigeons were able to discriminate between the tree and the non-tree slides. To confirm that this ability, impossible for any machine to match, was not somehow learnt through the long process of evolution and hardwired into the brain of the pigeons, another experiment was designed to check the discriminating ability of pigeons with respect to fish and non-fish scenes and once again the birds had no problem doing so. Over the years it has been shown that pigeons can also distinguish: (1) oak leaves from leaves of other trees, (ii) scenes with or without bodies of water, (iii) pictures showing a particular person from others with no people or different individuals.

Other examples of animal intelligence include mynah birds who can recognize trees or people in pictures, and signal their identification by vocal utterances---words---instead of pecking at buttons, and a parrot who can answer, vocally, questions about shapes and colours of objects, even those not seen before. The intelligence of higher animals, such as apes, elephants, and dolphins is even more remarkable.

Another recent summary of this research is that of Wasserman [5]: "[Experiments] support the conclusion that conceptualization is not unique to human beings. Neither having a human brain nor being able to use language is therefore a precondition for cognition... Complete understanding of neural activity and function must encompass the marvelous abilities of brains other than our own. If it is the business of brains to think and to learn, it should be the business of behavioral neuroscience to provide a full account of that thinking and learning in all animals---human and nonhuman alike."

An extremely important insight from experiments of animal intelligence is that one can attempt to define different gradations of cognitive function. It is obvious that animals are not as intelligent as humans; likewise, certain animals appear to be more intelligent than others. For example, pigeons did poorly at picking a pattern against two other identical ones, as in picking an A against two B's. This is a very simple task for humans.

Wasserman devised an experiment to show that pigeons could be induced to amalgamate two basic categories into one broader category not defined by any obvious perceptual features. The birds were trained to sort slides into two arbitrary categories, such as category of cars and people and the category of chairs and flowers. In the second part of this experiment, the pigeons were trained to reassign one of the stimulus classes in each category to a new response key. Next, they were tested to see whether they would generalize the reassignment to the stimulus class withheld during reassignment training. It was found that the average score was 87 percent in the case of stimuli that had been reassigned and 72 percent in the case of stimuli that had not been reassigned. This performance, exceeding the level of chance, indicated that perceptually disparate stimuli had amalgamated into a new category. A similar experiment was performed on preschool children. The children's score was 99 percent for stimuli that had been reassigned and 80 percent for stimuli that had not been reassigned. In other words, the children's performance was roughly comparable to that of pigeons. Clearly, the performance of adult humans at this task will be superior to that of children or pigeons.

Another interesting experiment related to the abstract concept of sameness. Pigeons were trained to distinguish between arrays composed of a single, repeating icon and arrays composed of 16 different icons chosen out of a library of 32 icons. During training each bird encountered only 16 of the 32 icons; during testing it was presented with arrays made up of the remaining 16 icons. The average score for training stimuli was 83 percent and the average score for testing stimuli was 71 percent. These figures show that an abstract concept not related to the actual associations learnt during training had been internalized by the pigeon.

Animal intelligence experiments suggest that one can speak of different styles of solving AI problems. Are the cognitive capabilities of pigeons limited because their style has fundamental limitations? It is possible that the relatively low scores on the sameness test for pigeons can be explained on the basis of wide variability in performance for individual pigeons and the unnatural conditions in which the experiments are performed. But is the cognitive style of all animals similar and the differences in their cognitive capabilities arise from the differences in the sizes of their mental hardware? Since current machines do not, and cannot, use inner representations, it is right to conclude that their performance can never match that of animals. Most importantly, the generalization achieved by pigeons and other nonhumans remains beyond the capability of machines.

A useful perspective on animal behavior is its recursive nature, or part-whole hierarchy. Considering this from the bottom up, animal societies have been viewed as *superorganisms*. For example, the ants in an ant colony may be compared to cells, their castes to tissues and organs, the queen and her drones to the generative system, and the exchange of liquid food amongst the colony members to the circulation of blood and lymph. Furthermore, corresponding to morphogenesis in organisms the ant colony has sociogenesis, which consists of the processes by which the individuals undergo changes in caste and behavior. Such recursion has been viewed all the way up to the earth itself seen as a living entity. Parenthetically, it may be asked whether the earth itself, as a living but unconscious organism, may not be viewed like the unconscious brain. Paralleling this recursion is the individual who can be viewed as a collection of several *agents* where these agents have sub-agents which are the sensory mechanisms and so on. But these agents are bound together and this binding defines consciousness.

**Holistic Processing and Quantum Models**

The quantum mechanical approach to the study of consciousness has an old history and the creators of quantum theory were amongst the first to suggest it. More recently, scholars have proposed specific quantum theoretic models of brain function, but there is no single model that has emerged as the favored one at this point. In my own work I have considered the connections between quantum theory and information arguing that brain's processing is organized in a hierarchy of languages: associative at the bottom, self-organizational in the middle, and quantum at the top. Neural learning is associative and it proceeds to create necessary structures to "measure" the stimulus-space; at the higher level of multiple agents the response is by reorganizing the grosser levels of the neural structure. Each cognitive agent is an abstract quantum system. The linkages amongst the agents are regulated by an appropriate quantum field. This allows the individual at the higher levels of abstraction to initiate cognition or action, leading to active behavior.

One striking success of the quantum models is that they provide a resolution to the determinism- free will problem. According to quantum theory, a system evolves causally until it is observed. The act of observation causes a break in the causal chain. This leads to the notion of a participatory universe [6]. Consciousness provides a break in the strict regime of causality. It would be reasonable to assume that this freedom is associated with all life. But its impact on the ongoing processes will depend on the entropy associated with the break in the causal chain.

Quantum theory defines knowledge in a relative sense. In the quantum world, it is meaningless to talk of an objective reality. Knowledge is a collection of the observations on the reductions of the wavefunction, brought about by measurements using different kinds of instrumentations.

The indeterminacy of quantum theory does not reside in the microworld alone. For example, Schrödinger's cat paradox shows how a microscopic uncertainty transforms into a microscopic uncertainty. Brain processes are not described completely by the neuron firings; one must, additionally, consider their higher order bindings, such as thoughts and abstract concepts, because they, in turn, have an influence on the neuron firings. A wavefunction describing the brain would then include variables for the higher order processes, such as abstract concepts as well. But such a definition will leave certain indeterminacy in our description.

**Reorganizing Signals**

Living systems are characterized by continual adaptive organization at various levels. The reorganization is a response to the complex of signal flows within the larger system. For example, the societies of ants or bees may be viewed as single superorganisms. Hormones and other chemical exchanges among the members of the colony determine the ontogenies of the individuals within the colony. More pronounced than this global exchange is the activity amongst the individuals in cliques or groups.

Paralleling trophallaxis is the exchange of neurotransmitters or electrical impulses within a neural network at one level, and the integration of sensory data, language, and ideas at other levels. An illustration of this is the adaptation of somatosensory cortex to differential inputs. The cortex enlarges its representation of particular fingers when they are stimulated, and it reduces its representation when the inputs are diminished, such as by limb deafferentation.

Adaptive organization may be a general feature of neural networks and of the neocortex in particular. Biological memory and learning within the cortex may be organized adaptively. While there are many ways of achieving this, nesting among neural networks within the cortex is a key principle in self-organization and adaptation. Nested distributed networks provide a means of orchestrating bottom-up and top-down regulation of complex neural processes operating within and between many levels of structure.

There may be at least two modes of signaling that are important within a nested arrangement of distributed networks. A fast system manifests itself as spatiotemporal patterns of activation among modules of neurons. These patterns flicker and encode correlations that are the signals of the networks within the cortex. They are analogous to the hormones and chemical exchanges of the ant or bee colonies in the example mentioned earlier. In the brain, the slow mode is mediated by such processes as protein phosphorylation and synaptic plasticity. They are the counterparts of individual ontogenies in the ant or bee colonies. The slow mode is intimately linked to learning and development (i.e., ontogeny), and experience with and adaptation to the environment affect both learning and memory.

By considering the question of adaptive organization in the cortex, our approach is in accordance with the ideas of Gibson [7] who has long argued that biological processing must be seen as an active process. We make the case that nesting among cortical structures provides a framework in which active reorganization can be efficiently and easily carried out. The processes are manifest by at least two different kinds of signaling, with the consequence that the cortex is viewed as a dynamic system at many levels, including the level of brain regions. Consequently, functional anatomy, including the realization of the homunculus in the motor and sensory regions, is also dynamic. The homunculus is an evolving, and not a static representation, in this view.

It is not known how appropriate associative modules come into play in response to a stimulus. This is an important open question in neural computing. The paradigm of "active" processing in the context of memory is usually treated in one of two ways. First, the processing may be pre-set. This is generally termed "supervised learning", and it is a powerful but limited form of active processing. A second type of processing does not involve an explicit teacher, and this mechanism is termed "unsupervised learning". It is sensitive to a number of constraints, including the structure and modulation of the network under consideration.

There are different ways that biological memory may be self-organizing, and in this section, we suggest that the nesting of distributed neural networks within the neocortex is a natural candidate for encoding and transducing memory. Nesting has interesting combinatorial and computational features, and the seemingly simplistic organization of nested neural networks may have profound computational properties. However, we do not claim that nesting is the only important feature for adaptive organization in neural systems.

The neocortex is a great expanse of neural tissue that makes up the bulk of the human brain. As in all other species, the human neocortex is made up of neural building blocks. At a rudimentary level, these blocks consist of columns oriented perpendicular to the surface of the cortex. These columns may be seen as organized in the most basic form as minicolumns of about 30 μm in diameter. The minicolumns are, in turn, organized into larger columns of approximately 500 - 1000 μm in diameter. It has been estimated that the human neocortex contains about 600 million minicolumns and about 600,000 larger columns. Columns are defined by ontogenetic and functional criteria, and there is evidence that columns in different brain regions coalesce into functional modules. Different regions of the brain have different architectonic properties, and subtle differences in anatomy are associated with differences in function.

Beginning with cortical minicolumns, progressive levels of cortical structure consist of columns, modules, regions and systems. It is assumed that these structures evolve and adapt through the lifespan. It is also assumed that the boundaries between the clusters are plastic: they change slowly due to synaptic modifications or, more rapidly, due to synchronous activity among adjacent clusters.

Results from the study of neural circuits controlling rhythmic behavior, such as feeding, locomotion, and respiration, show that the same network, through a process of "rewiring" can express different functional capabilities.

Superorganisms also have nested structures in terms of individuals who interact more with certain members than others. In the case of ants, the castes provide further "modular" structure. For the case of honeybees [8]: "[It is important to recognize] subsystems of communication, or cliques, in which the elements interact more frequently with each other than with other members of the communication system. In context, the dozen or so honeybee workers comprising the queen retinue certainly communicate more within their group (including the queen) than they do with the one or two hundred house bees receiving nectar loads from foragers returning from the field. The queen retinue forms one communication clique while the forager-receiver bees form another clique."

Another fundamental communication within the superorganism is the one that defines its constitution. This is a much slower process which can be seen, for example, when a queen ant founds her colony. The queen governs the process of caste morphogenesis [9]. Within the new colony, the queen, having just mated with her suitors and received more than 200 million sperm, shakes off her wings and digs a little nest in the ground, where she now is in a race with time to produce her worker offspring. She raises her first brood of workers by converting her body fat and muscles into energy. She must create a perfectly balanced work force that is the smallest possible in size, yet capable of successful foraging, so that the workers can bring food to her before she starves to death.

The queen produces the workers of the correct size for her initial survival and later, after the colony has started going, she produces a complement of workers of different sizes as well as soldier ants in order to have the right organization for the survival of the colony. When researchers have removed members of a specific caste from an ongoing colony, the queen compensates for this deficit by producing more members of that caste. The communication process behind this remarkable control is not known. The communication mechanisms of the ant or the honeybee superorganisms may be supposed to have analogs in the brain.

**Anomalous abilities and deficits amongst humans**

That cognitive ability cannot be viewed simply as a processing of sensory information by a central intelligence extraction system is confirmed by individuals with anomalous abilities. Idiot savants, or simply savants, who have serious developmental disability or major mental illness, perform spectacularly at certain tasks. Anomalous performance has been noted in the areas of mathematical and calendar calculations; music; art, including painting, drawing or sculpting; mechanical ability; prodigious memory (mnemonism); unusual sensory discrimination or "extrasensory" perception. The abilities of these savants and of mnemonists cannot be understood in the framework of a monolithic mind.

Oliver Sacks, in his book *The Man Who Mistook His Wife for a Hat* [10] describes two twenty-six year old twins, John and Michael, with IQs of sixty who are remarkable at calendrical calculations even though "they cannot do simple addition or subtraction with any accuracy, and cannot even comprehend what multiplication means." More impressive is their ability to factor numbers into primes since "primeness" is an abstract concept. Looking from an evolutionary perspective, it is hard to see that performing abstract numerical calculations related to primes would provide an advantage?

The architecture of the brain provides clues to the relationships amongst the agents, and this architecture is illuminated by examining deficits in function caused by injury.
One might expect aphasia to be accompanied by a general reduction in the capacity to talk, understand, read, write, as well as do mathematics and remember things. One might also suppose that the ability to read complex technical texts would be affected much more than the capacity to understand simple language and to follow commands.

In reality, the relationship between these capacities is very complex. In aphasia, many of these capacities, by themselves or in groups, can be destroyed or spared in isolation from the others. Historically, several capacities related to language have been examined. These include fluency in conversation, repetition, comprehension of spoken language, word-finding disability, and reading disturbances.

In alexia, the subject is able to write while unable to read; in alexia combined with agraphia, the subject is unable to write or read while retaining other language faculties; in acalculia, the subject has selective difficulty in dealing with numbers.

The complex manner in which these aphasias manifest establishes that language production is a very intricate process More specifically, it means that at least certain components of the language functioning process operate in a yes/no fashion.
These components include comprehension, production, repetition, and various abstract processes. But to view each as a separate module only tells half the story. There exist very subtle interrelationships between these capabilities which all come into operation in normal behavior.

**Blindsight and agnosia**

There are anecdotal accounts of blind people who can see sometime and deaf people who can likewise hear. Some brain damaged subjects cannot consciously see an object in front of them in certain places within their field of vision, yet when asked to guess if a light had flashed in their region of blindness, the subjects guess right at a probability much above that of chance.

Blindsight has been explained as being a process similar to that of implicit memory or it has been proposed that consciousness is a result of a dialog going on between different regions of the brain. When this dialog is disrupted, even if the sensory signals do reach the brain, the person will not be aware of the stimulus.

One may consider that the injury in the brain leading to blindsight causes the vision in the stricken field to become automatic. Then through retraining it might be possible to regain the conscious experience of the images in this field. In the holistic explanation, the conscious awareness is a correlate of the activity in a complex set of regions in the brain. No region can be considered to be producing the function by itself although damage to a specific region will lead to the loss of a corresponding function.

Agnosia is a failure of recognition that is not due to impairment of the sensory input or a general intellectual impairment. A visual agnosic patient will be unable to tell what he is looking at, although it can be demonstrated that the patient can see the object. Prosopagnosic patients are neither blind nor intellectually impaired; they can interpret facial expressions and they can recognize their friends and relations by name or voice. Yet they do not recognize specific faces, not even their own in a mirror!

Prosopagnosia may be regarded as the opposite of blindsight. In blindsight there is recognition without awareness, whereas in prosopagnosia there is awareness without recognition. But there is evidence that the two syndromes have underlying similarity. Electrodermal recordings show that the prosopagnosic responds to familiar faces although without awareness of this fact. It appears, therefore, that the patient is subconsciously registering the significance of the faces. Prosopagnosia may be suppressed under conditions of associative priming. Thus if the patient is shown the picture of some other face it may trigger a recognition.

**Split Brains and unification**

The two hemispheres of the brain are linked by the rich connections of the corpus callosum. The visual system is arranged so that each eye normally projects to both hemispheres. By cutting the optic-nerve crossing, the chiasm, the remaining fibers in the optic nerve transmit information to the hemisphere on the same side. Visual input to the left eye is sent only to the left hemisphere, and input to the right eye projects only to the right hemisphere. The visual areas also communicate through the corpus callosum. When these fibers are also severed, the patient is left with a split brain.

A classic experiment on cat with split brains was conducted by Ronald Myers and Roger Sperry [11]. They showed that cats with split brains did as well as normal cats when it came to learning the task of discriminating between a circle and a square in order to obtain a food reward, while wearing a patch on one eye. This showed that one half of the brain did as well at the task as both the halves in communication. When the patch was transferred to the other eye, the split-brain cats behaved different from the normal cats, indicating that their previous learning had not been completely transferred to the other half of the brain.

Experiments on split-brain human patients raised questions related to the nature and the seat of consciousness. For example, a patient with left-hemisphere speech does not know what his right hemisphere has seen through the right eye. The information in the right brain is unavailable to the left brain and vice versa. The left brain responds to the stimulus reaching it whereas the right brain responds to its own input. Each half brain learns, remembers, and carries out planned activities. It is as if each half brain works and functions outside the conscious realm of the other. Such behavior led Sperry to suggest that there are "two free wills in one cranial vault."

But there are other ways of looking at the situation. One may assume that the split-brain patient has lost conscious access to those cognitive functions which are regulated by the non-speech hemisphere. Or, one may say that nothing is changed as far as the awareness of the patient is considered and the cognitions of the right brain were linguistically isolated all along, even before the commissurotomy was performed. The procedure only disrupts the visual and other cognitive-processing pathways.

The patients themselves seem to support this second view. There seems to be no antagonism in the responses of the two hemispheres and the left hemisphere is able to fit the actions related to the information reaching the right hemisphere in a plausible theory. For example, consider the test where the word "pink" is flashed to the right hemisphere and the word "bottle" is flashed to the left. Several bottles of different colors and shapes are placed before the patient and he is asked to choose one. He immediately picks the pink bottle explaining that pink is a nice colour. Although the patient is not consciously aware of the right eye having seen the word "pink" he, nevertheless, "feels" that pink is the right choice for the occasion. In this sense, this behavior is very similar to that of blindsight patients.

The brain has many modular circuits that mediate different functions. Not all of these functions are part of conscious experience. When these modules related to conscious sensations get "crosswired," this leads to synesthesia. One would expect that similar joining of other cognitions is also possible. A deliberate method of achieving such a transition from many to one is a part of some meditative traditions.

It is significant that patients with disrupted brains never claim to have anything other than a unique awareness. The reductionists opine that consciousness is nothing but the activity in the brain but this is mere semantic play which sheds no light on the problem. If shared activity was all there was to consciousness, then this would have been destroyed or multiplied by commissurotomy. Split brains should then represent two minds just as in freak births with one trunk and two heads we do have two minds.

**Conclusions**

This article has considered evidence from physical and biological sciences to show how machines are deficient compared to biological systems at incorporating intelligence. To recapitulate the main points, machines fall short on two counts as compared to brains. Firstly, unlike brains, machines do not self-organize in a recursive manner. Secondly, machines are based on classical logic, whereas Nature's intelligence may depend on quantum mechanics.

Quantum mechanics provides us a means of obtaining information about a system associated with various attributes. A quantum state is a linear superposition of its component states. Since the amplitudes are complex numbers, a quantum system cannot be effectively simulated by the Monte Carlo method using random numbers. One cannot run a physical process if its probability amplitude is negative or complex!

The evidence from neuroscience that we reviewed showed how specific centers in the brain are dedicated to different cognitive tasks. But these centers do not merely do signal processing: each operates within the universe of its experience so that it is able to generalize individually. This generalization keeps up with new experience and is further related to other cognitive processes in the brain. It is in this manner that cognitive ability is holistic and irreducible to a mechanistic computing algorithm. Viewed differently, each agent is an apparatus that taps into the "universal field of consciousness." On the other hand, AI machines based on classical computing principles have a fixed universe of discourse so they are unable to adapt in a flexible manner to a changing universe. This is why they cannot match biological intelligence.

Quantum theory has the potential to provide understanding of certain biological processes not amenable to classical explanation. Take the protein-folding problem. Proteins are sequences of large number of amino acids. Once a sequence is established, the protein folds up rapidly into a highly specific three-dimensional structure that determines its function in the organism. It has been estimated that a fast computer applying plausible rules for protein folding would need $10^{127}$ years to find the final folded form for even a very short sequence of 100 amino acids [12]. Yet Nature solves this problem in a few seconds. Since quantum computing can be exponentially faster than conventional computing, it could very well be the explanation for Nature's speed [13]. The anomalous efficiency of other biological optimization processes may provide indirect evidence of underlying quantum processing if no classical explanation is forthcoming.

At an abstract level, if evolution of life has led to the emergence of mind, machines with minds must also emerge. Tipler and Barrow argue that man will create silicon machines with minds that will slowly spread all over the world, and the entire universe will eventually become a conscious machine [14].

In my view, if machines with consciousness are created, they would be living machines, that is, variations on life forms as we know them. Since the material world is not causally closed, and consciousness influences its evolution, matter and minds complement each other. At the level of the individual, even medical science that is strongly based on the machine paradigm is acknowledging the influence of mind on body [15].